\documentclass[sigconf]{acmart}

\usepackage{booktabs} 
\usepackage{url}

\usepackage{amsmath}	
\usepackage{amssymb}
\usepackage{graphicx}
\usepackage{hyperref}
\usepackage{wrapfig}
\usepackage{cleveref}
\renewcommand{\vec}[1]{\mathbf{#1}}

\usepackage{listings}
\usepackage{color}

\definecolor{dkgreen}{rgb}{0,0.6,0}
\definecolor{gray}{rgb}{0.5,0.5,0.5}
\definecolor{mauve}{rgb}{0.58,0,0.82}

\begin{document}

\copyrightyear{2018}
\acmYear{2018}
\setcopyright{acmcopyright}
\acmConference[ICMLC 2018]{2018 10th International Conference on Machine Learning and Computing}{February 26--28, 2018}{Macau, China}
\acmBooktitle{ICMLC 2018: 2018 10th International Conference on Machine Learning and Computing, February 26--28, 2018, Macau, China}
\acmPrice{15.00}
\acmDOI{10.1145/3195106.3195117}
\acmISBN{978-1-4503-6353-2/18/02}

\title[GRU-SVM]{A Neural Network Architecture Combining Gated Recurrent Unit (GRU) and Support Vector Machine (SVM) for Intrusion Detection in Network Traffic Data}

\author{Abien Fred M. Agarap}
\email{abienfred.agarap@gmail.com}

\begin{abstract}
Gated Recurrent Unit (GRU) is a recently-developed variation of the long short-term memory (LSTM) unit, both of which are variants of recurrent neural network (RNN). Through empirical evidence, both models have been proven to be effective in a wide variety of machine learning tasks such as natural language processing\cite{wen2015semantically}, speech recognition\cite{chorowski2015attention}, and text classification\cite{yang2016hierarchical}. Conventionally, like most neural networks, both of the aforementioned RNN variants employ the Softmax function as its final output layer for its prediction, and the cross-entropy function for computing its loss. In this paper, we present an amendment to this norm by introducing linear support vector machine (SVM) as the replacement for Softmax in the final output layer of a GRU model. Furthermore, the cross-entropy function shall be replaced with a margin-based function. While there have been similar studies\cite{Alalshekmubarak, Tang}, this proposal is primarily intended for binary classification on intrusion detection using the 2013 network traffic data from the honeypot systems of Kyoto University. Results show that the GRU-SVM model performs relatively higher than the conventional GRU-Softmax model. The proposed model reached a training accuracy of $\approx$81.54\% and a testing accuracy of $\approx$84.15\%, while the latter was able to reach a training accuracy of $\approx$63.07\% and a testing accuracy of $\approx$70.75\%. In addition, the juxtaposition of these two final output layers indicate that the SVM would outperform Softmax in prediction time - a theoretical implication which was supported by the actual training and testing time in the study.
\end{abstract}



\maketitle

\section{Introduction}

By 2019, the cost to the global economy due to cybercrime is projected to reach \$2 trillion\cite{Juniper}. Among the contributory felonies to cybercrime is \textit{intrusions}, which is defined as illegal or unauthorized use of a network or a system by attackers\cite{Ghosh}. An intrusion detection system (IDS) is used to identify the said malicious activity\cite{Ghosh}. The most common method used for uncovering intrusions is the analysis of user activities\cite{Ghosh, Kumar, Mukkamala}. However, the aforementioned method is laborious when done manually, since the data of user activities is massive in nature\cite{Frank, Lincoln}. To simplify the problem, automation through machine learning must be done.\\
\indent	A study by Mukkamala, Janoski, \& Sung (2002)\cite{Mukkamala} shows how \textit{support vector machine} (SVM) and \textit{artificial neural network} (ANN) can be used to accomplish the said task. In machine learning, SVM separates two classes of data points using a hyperplane\cite{Cortes}. On the other hand, an ANN is a computational model that represents the human brain, and shows information is passed from a neuron to another\cite{Negnevitsky}.\\
\indent	An approach combining ANN and SVM was proposed by Alalshekmubarak \& Smith\cite{Alalshekmubarak}, for time-series classification. Specifically, they combined \textit{echo state network} (ESN, a variant of recurrent neural network or RNN) and SVM. This research presents a modified version of the aforementioned proposal, and use it for intrusion detection. The proposed model will use \textit{recurrent neural network} (RNNs) with \textit{gated recurrent units} (GRUs) in place of ESN. RNNs are used for analyzing and/or predicting sequential data, making it a viable candidate for intrusion detection\cite{Negnevitsky}, since network traffic data is sequential in nature.

\section{Methodology}
\subsection{Machine Intelligence Library}
Google TensorFlow\cite{tensorflow2015-whitepaper} was used to implement the neural network models in this study -- both the proposed and its comparator. 

\subsection{The Dataset}
The 2013 Kyoto University honeypot systems' network traffic data\cite{Song} was used in this study. It has 24 statistical features\cite{Song}; (1) 14 features from the KDD Cup 1999 dataset\cite{Stolfo}, and (2) 10 additional features, which according to Song, Takakura, \& Okabe (2006)\cite{Song}, might be pivotal in a more effective investigation on intrusion detection. Only 22 dataset features were used in the study.

\subsection{Data Preprocessing}

For the experiment, only 25\% of the whole 16.2 GB network traffic dataset was used, i.e. $\approx$4.1 GB of data (from January 1, 2013 to June 1, 2013). Before using the dataset for the experiment, it was normalized first -- standardization (for continuous data, see Eq. \ref{z-formula}) and indexing (for categorical data), then it was binned (discretized).

\begin{equation}\label{z-formula}
z = \dfrac{X - \mu}{\sigma}
\end{equation}

\indent	where $X$ is the feature value to be standardized, $\mu$ is the mean value of the given feature, and $\sigma$ is its standard deviation. But for efficiency, the \texttt{StandardScaler().fit\_transform()} function of Scikit-learn\cite{scikit-learn} was used for the data standardization in this study.\\
\indent	For indexing, the categories were mapped to $[0, n-1]$ using the \texttt{LabelEncoder().fit\_transform()} function of Scikit-learn\cite{scikit-learn}.
\indent	After dataset normalization, the continuous features were binned (decile binning, a discretization/quantization technique). This was done by getting the $10^{th}$, $20^{th}$, ..., $90^{th}$, and $100^{th}$ quantile of the features, and their indices served as their bin number. This process was done using the \texttt{qcut()} function of \texttt{pandas}\cite{mckinney-proc-scipy-2010}. Binning reduces the required computational cost, and improves the classification performance on the dataset\cite{lustgarten2008improving}. Lastly, the features were one-hot encoded, making it ready for use by the models.

\subsection{The GRU-SVM Neural Network Architecture}
\begin{figure}[!htb]
\minipage{0.5\textwidth}
\centering
	\includegraphics[width=\linewidth]{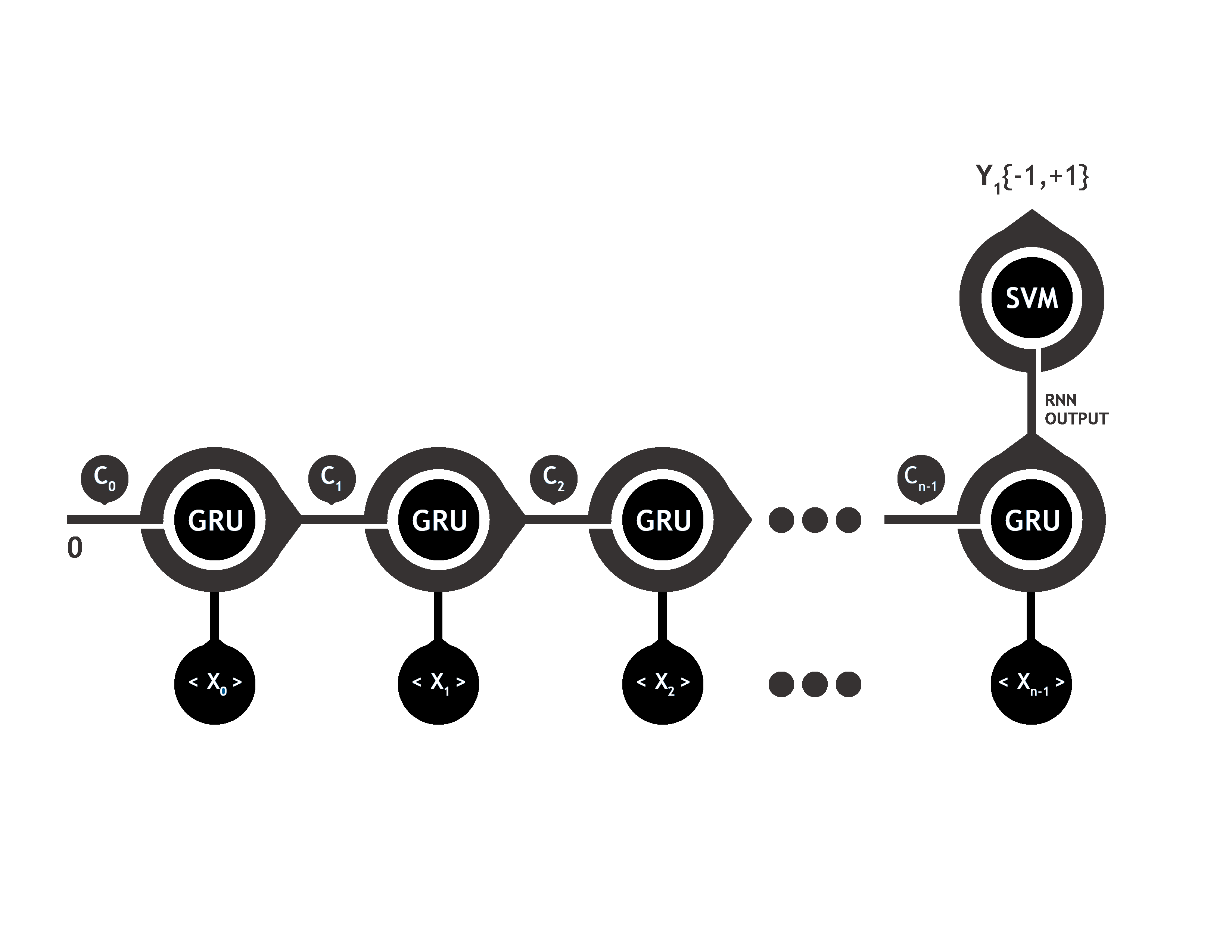}
	\caption{The proposed GRU-SVM architecture model, with $n-1$ GRU unit inputs, and SVM as its classifier.}
	\label{fig:proposed-model}
\endminipage\hfill
\end{figure}

Similar to the work of Alalshekmubarak \& Smith (2013)\cite{Alalshekmubarak} and Tang (2013)\cite{Tang}, the present paper proposes to use SVM as the classifier in a neural network architecture. Specifically, a Gated Recurrent Unit (GRU) RNN (see Figure \ref{fig:proposed-model}).

\indent	For this study, there were 21 features used as the model input. Then, the parameters are learned through the gating mechanism of GRU\cite{Cho} (\Crefrange{z-gate}{new-value}).
\begin{equation}\label{z-gate}
z	=	\sigma(\vec{W}_{z} \cdot [h_{t - 1}, x_{t}])
\end{equation}
\begin{equation}\label{r-gate}
r	=	\sigma(\vec{W}_{r} \cdot [h_{t - 1}, x_{t}])
\end{equation}
\begin{equation}\label{candidate-value}
\tilde{h}_{t}	=	tanh(\vec{W} \cdot [r_{t} * h_{t - 1}, x_{t}])
\end{equation}
\begin{equation}\label{new-value}
h_{t}	=	(1 - z_{t}) * h_{t - 1} + z_{t} * \tilde{h}_{t}
\end{equation}

But with the introduction of SVM as its final layer, the parameters are also learned by optimizing the objective function of SVM (see Eq. \ref{l1-svm}). Then, instead of measuring the network loss using cross-entropy function, the GRU-SVM model will use the loss function of SVM (Eq. \ref{l1-svm}).

\begin{equation}\label{l1-svm}
min \dfrac{1}{2}\|\vec{w}\|_{1}^{2} + C \sum_{i = 1}^{n} max(0, 1 - y_{i}'(\vec{w}^{T}\vec{x}_{i}+b_{i}))
\end{equation}

Eq. \ref{l1-svm} is known as the unconstrained optimization problem of L1-SVM. However, it is not differentiable. On the contrary, its variation, known as the L2-SVM is differentiable and is more stable\cite{Tang} than the L1-SVM:
\begin{equation}\label{l2-svm}
min \dfrac{1}{2}\|\vec{w}\|_{2}^{2} + C \sum_{i = 1}^{n} max(0, 1 - y_{i}'(\vec{w}^{T}\vec{x}_{i}+b_{i}))^{2}
\end{equation}

The L2-SVM was used for the proposed GRU-SVM architecture. As for the prediction, the decision function $f(x) = sign(\vec{w}\vec{x}+b)$ produces a score vector for each classes. So, to get the predicted class label $y$ of a data $x$, the $argmax$ function is used:
\begin{align*}
predicted\_class &= argmax(sign(\vec{w}\vec{x} + b))
\end{align*}

The $argmax$ function will return the index of the highest score across the vector of the predicted classes.\\
\indent	The proposed GRU-SVM model may be summarized as follows:
\begin{enumerate}
\item	Input the dataset features $\{\vec{x}_{i}\ |\ \vec{x}_{i} \in \mathbb{R}^{m}\}$ to the GRU model.
\item	Initialize the learning parameters \texttt{weights} and \texttt{biases} with arbitrary values (they will be adjusted through training).
\item	The cell states of GRU are computed based on the input features $\vec{x}_{i}$, and its learning parameters values.
\item	At the last time step, the prediction of the model is computed using the decision function of SVM: $f(x) = sign(\vec{w}\vec{x} + b)$.
\item	The loss of the neural network is computed using Eq. \ref{l2-svm}.
\item	An optimization algorithm is used for loss minimization (for this study, the Adam\cite{Kingma} optimizer was used). Optimization adjusts the \texttt{weights} and \texttt{biases} based on the computed loss.
\item	This process is repeated until the neural network reaches the desired accuracy or the highest accuracy possible. Afterwards, the trained model can be used for binary classification on a given data.
\end{enumerate}

The program implementation of the proposed GRU-SVM model is available at https://github.com/AFAgarap/gru-svm.

\subsection{Data Analysis}
The effectiveness of the proposed GRU-SVM model was measured through the two phases of the experiment: (1) training phase, and (2) test phase. Along with the proposed model, the conventional GRU-Softmax was also trained and tested on the same dataset.\\
\indent	The first phase of the experiment utilized 80\% of total data points ($\approx$3.2 GB, or 14, 856, 316 lines of network traffic log) from the 25\% of the dataset. After normalization and binning, it was revealed through a high-level inspection that a duplication occurred. Using the \texttt{DataFrame.drop\_duplicates()} of pandas\cite{mckinney-proc-scipy-2010}, the 14, 856, 316-line data dropped down to 1, 898, 322 lines ($\approx$40MB).\\
\indent	The second phase of the experiment was the evaluation of the two trained models using 20\% of total data points from the 25\% of the dataset. The testing dataset also experienced a drastic shrinkage in size -- from 3, 714, 078 lines to 420, 759 lines ($\approx$9 MB). \\
\indent	The parameters for the experiments are the following: (1) Accuracy, (2) Epochs, (3) Loss, (4) Run time, (5) Number of data points, (6) Number of false positives, (7) Number of false negatives. These parameters are based on the ones considered by Mukkamala, Janoski, \& Sung (2002)\cite{Mukkamala} in their study where they compared SVM and a feed-forward neural network for intrusion detection. Lastly, the statistical measures for binary classification were measured (true positive rate, true negative rate, false positive rate, and false negative rate).

\section{Results}

All experiments in this study were conducted on a laptop computer with Intel Core(TM) i5-6300HQ CPU @ 2.30GHz x 4, 16GB of DDR3 RAM, and NVIDIA GeForce GTX 960M 4GB DDR5 GPU. The hyperparameters used for both models were assigned by hand, and not through hyper-parameter optimization/tuning (see Table \ref{table: hyperparameters}).

\begin{table}\centering
\caption{Hyper-parameters used in both neural networks.}
		\begin{tabular}{ccc}
		\toprule
		Hyper-parameters & GRU-SVM & GRU-Softmax \\
		\midrule
		Batch Size & 256 & 256\\
		Cell Size & 256 & 256\\
		Dropout Rate & 0.85 & 0.8\\
		Epochs & 5 & 5\\
		Learning Rate & 1e-5 & 1e-6\\
		SVM C & 0.5 & N/A \\
		\bottomrule
		\end{tabular}\\
		\label{table: hyperparameters}
\end{table}

Both models were trained on 1,898,240 lines of network traffic data for 5 epochs. Afterwards, the trained models were tested to classify 420,608 lines of network traffic data for 5 epochs. Only the specified number of lines of network traffic data were used for the experiments as those are the values that are divisble by the batch size of 256. The class distribution of both training and testing dataset is specified in Table \ref{class-distribution}.\\
\indent	The experiment results are summarized in Table \ref{table: summary-results}. Although the loss for both models were recorded, it will not be a topic of further discussion as they are not comparable since they are in different scales. Meanwhile, Tables \ref{statistical-measures-training} \& \ref{statistical-measures-testing} show the statistical measures for binary classification by the models during training and testing.

\begin{table}
\centering
\caption{Class distribution of training and testing dataset.}
\label{class-distribution}
	\begin{tabular}{ccc}
	\toprule
	Class & Training data & Testing data \\
	\midrule
	Normal & 794,512 & 157,914\\
	Intrusion detected & 1,103,728 & 262,694\\
	\bottomrule
	\end{tabular}
\end{table}
\begin{table}
\centering
\caption{Summary of experiment results on both GRU-SVM and GRU-Softmax models.}
		\begin{tabular}{ccc}
		\toprule
		Parameter & GRU-SVM & GRU-Softmax \\
		\midrule
		No. of data points -- Training & 1,898,240 & 1,898,240\\
		No. of data points -- Testing & 420,608 & 420,608 \\
		Epochs & 5 & 5\\
		Accuracy -- Training & $\approx$81.54\% & $\approx$63.07\%\\
		Accuracy -- Testing & $\approx$84.15\% & $\approx$70.75\%\\
		Loss -- Training &  $\approx$131.21& $\approx$0.62142\\
		Loss -- Testing & $\approx$129.62 & $\approx$0.62518\\
		Run time -- Training & $\approx$16.72mins & $\approx$17.18mins\\
		Run time -- Testing & $\approx$1.37mins & $\approx$1.67mins \\
		No. of false positives -- Training & 889,327 & 3,017,548\\
		No. of false positives -- Testing & 192,635 & 32,255\\
		No. of false negatives -- Training & 862,419 & 487,175\\
		No. of false negatives -- Testing & 140,535& 582,105\\
		\bottomrule
		\end{tabular}\\
		\label{table: summary-results}
\end{table}

\begin{table}
\centering
\caption{Statistical measures on binary classification: Training performance of the GRU-SVM and GRU-Softmax models.}
	\begin{tabular}{ccc}
	\toprule
		Parameter & GRU-SVM & GRU-Softmax\\
	\midrule
		True positive rate & $\approx$84.3726\% & $\approx$91.1721\% \\
		True negative rate & $\approx$77.6132\% & $\approx$24.0402\% \\
		False positive rate & $\approx$22.3867\% & $\approx$75.9597\% \\
		False negative rate & $\approx$15.6273\% & $\approx$8.82781\% \\
	\bottomrule
	\end{tabular}\\
	\label{statistical-measures-training}
\end{table}

\begin{table}
\centering
\caption{Statistical measures on binary classification: Testing performance of the GRU-SVM and GRU-Softmax models.}
	\begin{tabular}{ccc}
	\toprule
		Parameter & GRU-SVM & GRU-Softmax\\
	\midrule
		True positive rate & $\approx$89.3005\% & $\approx$55.6819\% \\
		True negative rate & $\approx$75.6025\% & $\approx$95.9149\% \\
		False positive rate & $\approx$10.6995\% & $\approx$4.08513\% \\
		False negative rate & $\approx$24.3975\% & $\approx$44.3181\% \\
	\bottomrule
	\end{tabular}\\
	\label{statistical-measures-testing}
\end{table}

\begin{figure}
\minipage{0.5\textwidth}\centering
	\includegraphics[width=\linewidth]{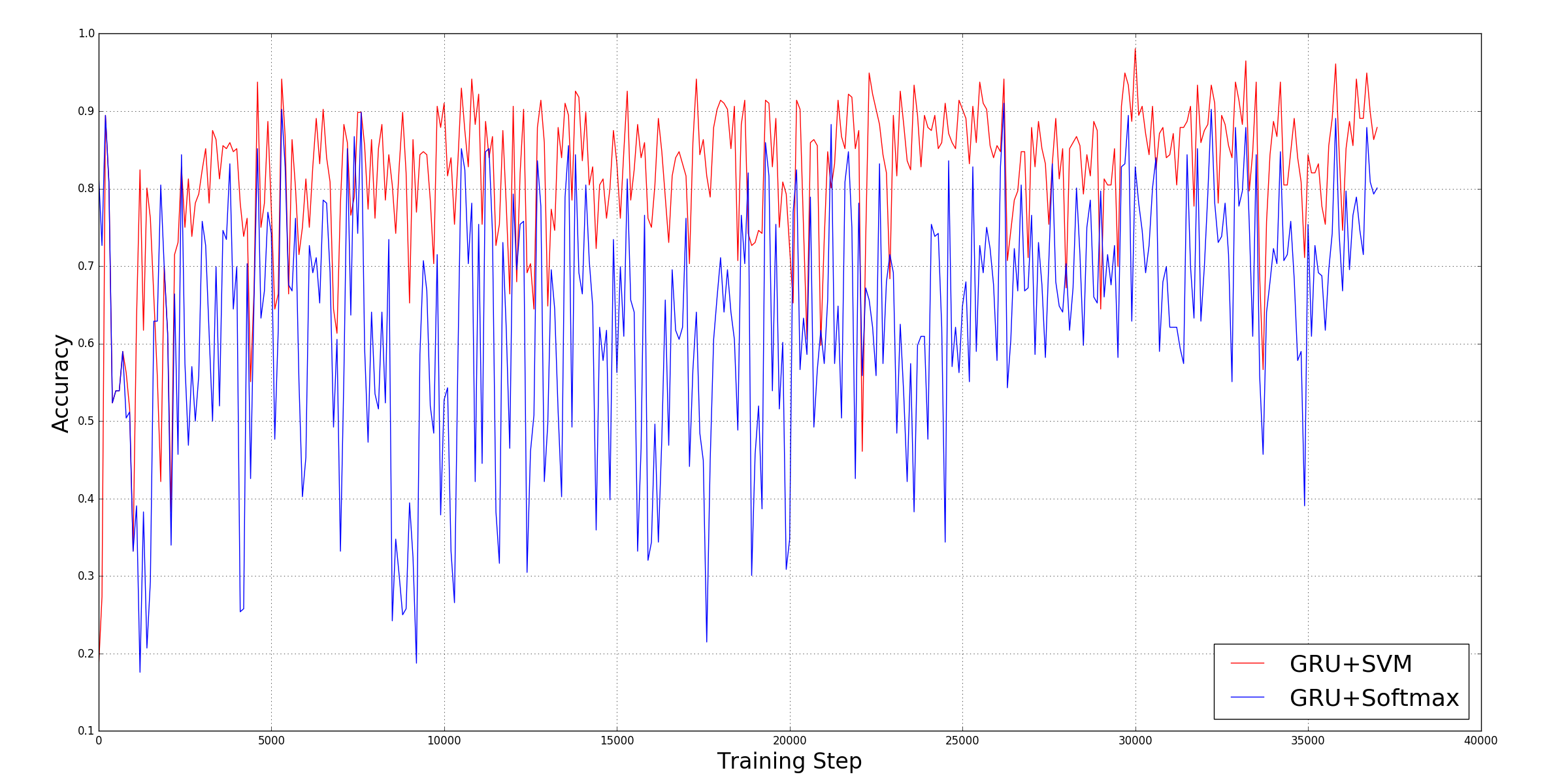}
	\caption{Training accuracy of the proposed GRU-SVM model, and the conventional GRU-Softmax model.}
	\label{training-accuracy}
\endminipage\hfill
\end{figure}

\begin{figure}
\minipage{0.5\textwidth}\centering
	\includegraphics[width=\linewidth]{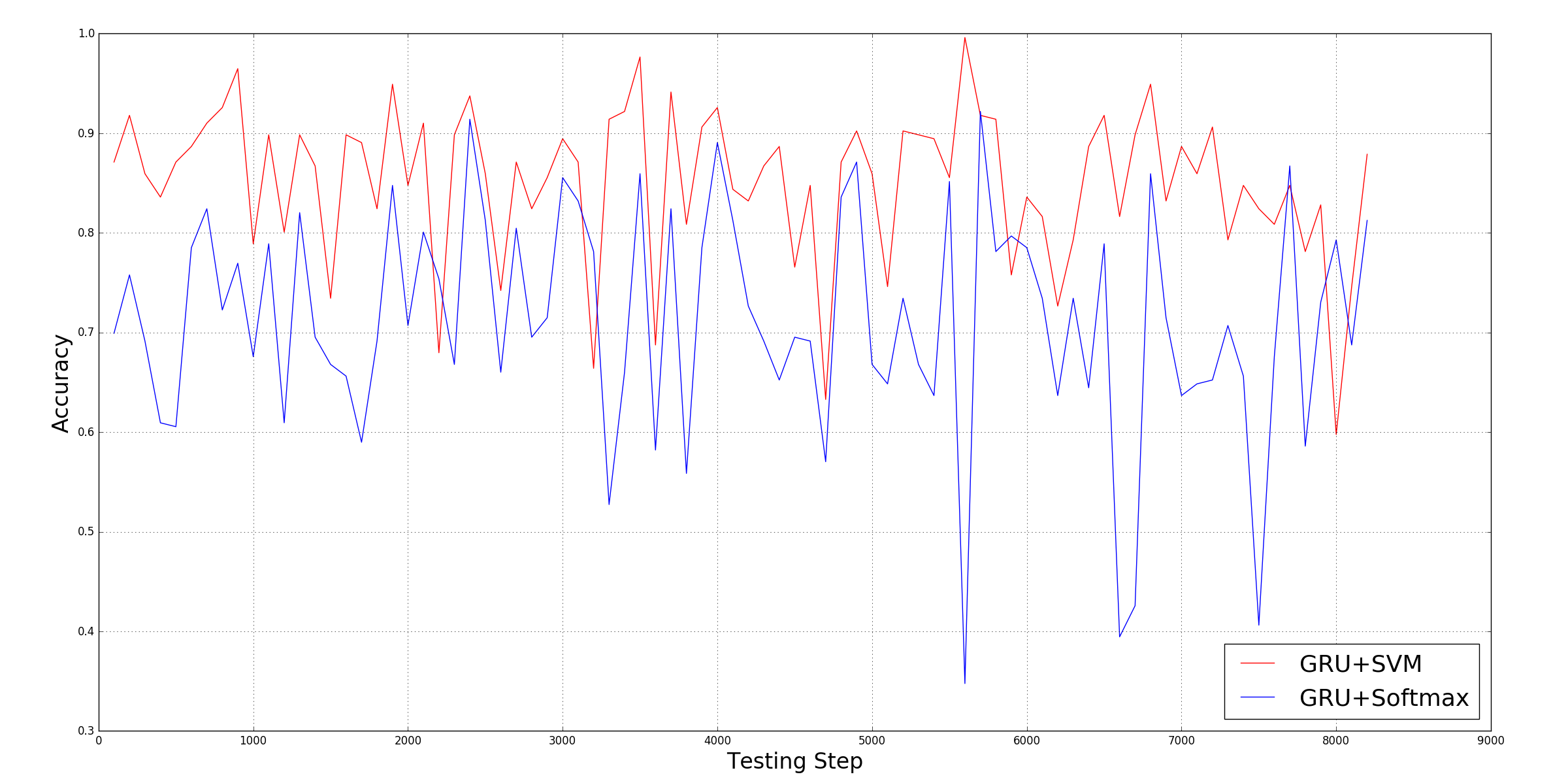}
	\caption{Testing accuracy of the proposed GRU-SVM model, and the conventional GRU-Softmax model.}
	\label{testing-accuracy}
\endminipage\hfill
\end{figure}

Figure \ref{training-accuracy} shows that for 5 epochs on the 1,898,240-line network traffic data (a total exposure of 9,491,200 to the training dataset), the GRU-SVM model was able to finish its training in 16 minutes and 43 seconds. On the other hand, the GRU-Softmax model finished its training in 17 minutes and 11 seconds.

Figure \ref{testing-accuracy} shows that for 5 epochs on the 420,608-line network traffic data (a total test prediction of 2,103,040), the GRU-SVM model was able to finish its testing in 1 minute and 22 seconds. On the other hand, the GRU-Softmax model finished its testing in 1 minute and 40 seconds.

\section{Discussion}

The empirical evidence presented in this paper suggests that SVM outperforms Softmax function in terms of prediction accuracy, when used as the final output layer in a neural network. This finding corroborates the claims by Alalshekmubarak \& Smith (2013)\cite{Alalshekmubarak} and Tang (2013)\cite{Tang}, and supports the claim that SVM is a more practical approach than Softmax for binary classification. Not only did the GRU-SVM model outperform the GRU-Softmax in terms of prediction accuracy, but it also outperformed the conventional model in terms of training time and testing time. Thus, supporting the theoretical implication as per the respective algorithm complexities of each classifier.\\
\indent	The reported training accuracy of $\approx$81.54\% and testing accuracy of $\approx$84.15\% posits that the GRU-SVM model has a relatively stronger predictive performance than the GRU-Softmax model (with training accuracy of $\approx$63.07\% and testing accuracy of $\approx$70.75\%). Hence, we propose a theory to explain the relatively lower performance of Softmax compared to SVM in this particular scenario. First, SVM was designed primarily for binary classification\cite{Cortes}, while Softmax is best-fit for multinomial classification\cite{Karpathy}. Building on the premise, SVM does not care about the individual scores of the classes it predicts, it only requires its margins to be satisfied\cite{Karpathy}. On the contrary, the Softmax function will always find a way to improve its predicted probability distribution by ensuring that the correct class has the higher/highest probability, and the incorrect classes have the lower probability. This behavior of the Softmax function is exemplary, but excessive for a problem like binary classification. Given that the sigmoid $\sigma$ function is a special case of Softmax (see Eq. \ref{sigmoid-softmax}-\ref{softmax-binary}), we can refer to its graph as to how it classifies a network output.
\begin{align}\label{sigmoid-softmax}
\sigma(y)	&=	\dfrac{1}{1 + e^{-y}}	=	\dfrac{1}{1 + \dfrac{1}{e^{y}}}	=	\dfrac{1}{\dfrac{e^{y} + 1}{e^{y}}}	=	\dfrac{e^{y}}{1 + e^{y}}	=	\dfrac{e^{y}}{e^{0}+e^{y}}
\end{align}
\begin{align}\label{softmax-binary}
softmax(y)	&=	\dfrac{e^{y_{i}}}{\sum_{i=0}^{n=1} e^{y_{i}}}	=	\dfrac{e^{y_{i}}}{e^{y_{0}} + e^{y_{1}}}
\end{align}

\begin{figure}
\minipage{0.5\textwidth}
\centering
	\includegraphics[width=0.95\linewidth]{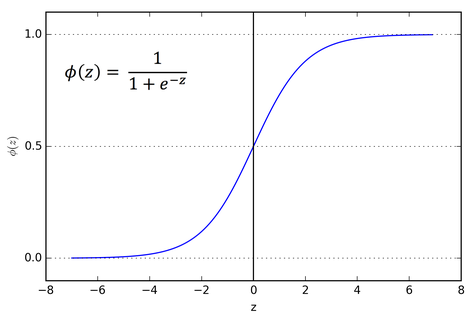}
	\caption{Image from \cite{45793856}. Graph of a sigmoid $\sigma$ function.}
	\label{sigmoid-graph}
\endminipage\hfill
\end{figure}

\indent	It can be inferred from the graph of sigmoid $\sigma$ function (see Figure \ref{sigmoid-graph}) that $y$ values tend to respond less to changes in $x$. In other words, the gradients would be small, which gives rise to the ``vanishing gradients'' problem. Indeed, one of the problems being solved by LSTM, and consequently, by its variants such as GRU\cite{Cho, hochreiter1997long}. This behavior defeats the purpose of GRU and LSTM solving the problems of a traditional RNN. We posit that this is the cause of misclassifications by the GRU-Softmax model.\\
\indent	The said erroneous manner of the GRU-Softmax model reflects as a favor for the GRU-SVM model. But the comparison of the exhibited predictive accuracies of both models is not the only reason for the practicality in choosing SVM over Softmax in this case. The amount of training time and testing time were also considered. As their computational complexities suggest, SVM has the upper hand over Softmax. This is because the algorithm complexity of the predictor function in SVM is only $O(1)$. On the other hand, the predictor function of Softmax has an algorithm complexity of $O(n)$. As results have shown, the GRU-SVM model also outperformed the GRU-Softmax model in both training time and testing time. Thus, it corroborates the respective algorithm complexities of the classifiers.\\

\section{Conclusion and Recommendation}

We proposed an amendment to the architecture of GRU RNN by using SVM as its final output layer in a binary/non-probabilistic classification task. This amendment was seen as viable for the fast prediction time of SVM compared to Softmax. To test the model, we conducted an experiment comparing it with the established GRU-Softmax model. Consequently, the empirical data attests to the effectiveness of the proposed GRU-SVM model over its comparator in terms of predictive accuracy, and training and testing time.\\
\indent	Further work must be done to validate the effectiveness of the proposed GRU-SVM model in other binary classification tasks. Extended study on the proposed model for a faster multinomial classification would prove to be prolific as well. Lastly, the theory presented to explain the relatively low performance of the Softmax function as a binary classifier might be a pre-cursor to further studies.

\section{Acknowledgment}
An appreciation to the open source community (Cross Validated, GitHub, Stack Overflow) for the virtually infinite source of information and knowledge; to the Kyoto University for their intrusion detection dataset from their honeypot system.

\bibliographystyle{ACM-Reference-Format}
\bibliography{paper} 

\end{document}